\documentclass[usenames,dvipsnames]{iopjournal}
\usepackage{hyphenat}
\usepackage{amssymb}
\usepackage{amsmath}
\usepackage{hyperref}
\usepackage{microtype}
\usepackage{booktabs}
\usepackage{multirow}
\usepackage[dvipsnames]{xcolor}

\begin{document}

\articletype{Article type} 

\title{{Text-Guided Diffusion-Based Adversarial Attacks on Chest X-Ray Images}}

\author{Basudha Pal$^1$\orcid{0009-0009-0920-8565}}, {Arjun Narayanan$^2$\orcid{0009-0004-3227-2569}}, {Neha Ajith$^{2}$\orcid{0009-0003-8795-9299}}, {Vikas R Bhat$^{2}$\orcid{0000-0001-9367-4925}} and {Muhammad Umair$^{4, 5}$\orcid{0000-0001-6113-8335}}

\affil{$^1$Department of Electrical and Computer Engineering, The Johns Hopkins University, Baltimore, 21218, MD, USA}

\affil{$^2$Manipal Institute of Technology, Manipal Academy of Higher Education, Manipal, 576104, Karnataka, India}

\affil{$^3$Department of Biomedical Engineering, The Johns Hopkins University, Baltimore, 21218, MD, USA}

\affil{$^4$The Russell H. Morgan Department of Radiology and Radiological Sciences, The Johns Hopkins Hospital, Baltimore, 21287, MD, USA}

\affil{$^5$Department of Radiology, Columbia University Irving Medical Center, New York, 10032, NY, USA}

\affil{$^*$Author to whom any correspondence should be addressed.}

\email{bpal5@jhu.edu}


\begin{abstract}
As artificial intelligence is increasingly integrated into chest X-ray (CXR) interpretation, triage, and clinical decision support, understanding its vulnerability to adversarial manipulation is critical for safe deployment. Existing robustness evaluations, however, predominantly rely on pixel-space attacks that introduce numerically constrained perturbations but may not represent plausible radiographic variation. This limitation is particularly important in multi-disease CXR classification, where models simultaneously evaluate multiple overlapping pathologies and adversarial failures may alter several diagnostic predictions. We propose a text-guided diffusion-based adversarial framework that optimizes learnable text conditioning while keeping the diffusion generator and target classifier frozen, enabling adversarial generation through a learned image prior rather than direct pixel manipulation. We evaluate the framework across multiple classifier architectures in both binary atelectasis and multi-disease CXR classification and compare it with FGSM, PGD, and Carlini--Wagner attacks. Our approach consistently produced the greatest degradation in classifier performance, reducing AUROC to 0.3885--0.5646 in binary classification and 0.4441--0.4878 in the multi-disease setting, while achieving superior image fidelity (SSIM 0.9080, LPIPS 0.1670, FID 51.23). Importantly, clinician interpretation remained unchanged for 95.9\% of binary and 73.8\% of multi-disease adversarial images despite substantial changes in model predictions. These findings reveal a clinically important discrepancy between human and machine interpretation and demonstrate the need to extend medical AI robustness evaluation beyond conventional pixel-space attacks toward generative threat models that can expose failures under visually and clinically plausible image variations.
\end{abstract}

\section{Introduction}

Chest radiography is one of the most widely used diagnostic imaging modalities and plays a central role in the detection, monitoring, and management of thoracic disease. The availability of large-scale public datasets, including ChestXray8 and ChestXray14, has accelerated the development of deep learning systems for automated chest X-ray (CXR) interpretation \cite{Wang2017}. Convolutional neural networks and, more recently, large vision and multimodal models have demonstrated strong performance across thoracic disease classification tasks, with early systems such as \textit{CheXNet} reporting performance comparable to expert radiologists for pneumonia detection \cite{Rajpurkar2017}. These advances have increased interest in AI for clinical decision support, triage, worklist prioritization, and secondary interpretation. The computational imaging methods in biomedical applications have similarly presented the value of mathematical modelling and reconstruction for obtaining clinically meaningful information from physiological recordings\cite{Bhat2026}.
Clinical deployment, however, requires more than high predictive performance under standard test conditions. Deep neural networks may rely on complex image features that differ from those used by human clinicians, allowing small input modifications to produce disproportionately large changes in prediction. Medical image classifiers are consequently vulnerable to carefully constructed adversarial examples \cite{Li2020}, making both white-box and black-box attacks an important consideration for safe deployment \cite{Alotaibi2025}. This vulnerability has also been demonstrated specifically in CXR classification, where attacks can substantially degrade diagnostic performance and susceptibility varies across architectures \cite{Lung2024}. CXR models may be particularly sensitive because disease prediction often depends on subtle localized findings while highly parameterized networks can exploit visual signals that are not clinically meaningful \cite{Lee2025}. Existing reviews report substantial vulnerability across CXR disease-classification tasks \cite{Sorin2023} and, more broadly, across radiological classification, segmentation, and report-generation systems \cite{Dietrich2025}. Adversarial perturbations may additionally transfer between CXR architectures, extending the threat beyond a single known target model \cite{Abou2024}.

Despite the rapid expansion of AI-based CXR interpretation, adversarial attack research in this setting remains comparatively limited, and existing evaluations are dominated by attacks defined directly in image space. Early CXR studies demonstrated vulnerability to the Fast Gradient Sign Method (FGSM) \cite{FGSM2015,Taghanaki2018}, while stronger iterative or optimization-based approaches such as Projected Gradient Descent (PGD) \cite{PGD2019}, Carlini--Wagner (C\&W) \cite{C&W2017}, and the Basic Iterative Method (BIM) \cite{Ma2021} have subsequently been applied to medical image classifiers. Comparative CXR evaluations consistently demonstrate vulnerability to such attacks \cite{abc1,abc2}, including residual susceptibility following adversarial training \cite{abc3}, while universal adversarial perturbations show that attack patterns can generalize across images \cite{abc4}. As CXR AI moves toward increasingly capable foundation and multimodal models and becomes more closely integrated with clinical workflows, evaluating robustness beyond these established attack formulations is therefore timely.

A central limitation of conventional attacks is the threat model they evaluate. FGSM, PGD, and C\&W primarily ask whether a decision boundary can be crossed by directly manipulating image pixels while constraining the numerical magnitude of the perturbation, typically under an $\ell_p$ norm. Yet a small pixel-space distance does not imply a clinically plausible radiographic variation. Gradient-based optimization can introduce high-frequency patterns, altered image statistics, or other artifacts that are mathematically small but unrelated to anatomical, physiological, or acquisition-related variability \cite{Alotaibi2025}. This distinction is especially important in medical imaging, where anatomical coherence and preservation of diagnostically meaningful structures are central to whether an altered image remains clinically valid.

This motivates a stronger question for adversarial robustness: \textit{can a classifier be driven to failure while the resulting image remains a plausible realization of the underlying radiographic distribution?} Such a threat model tests more than local sensitivity to additive noise. If an AI prediction changes substantially under a visually and clinically coherent modification while the human interpretation remains stable, the failure suggests that the classifier may depend on fragile or non-clinical representations rather than robust pathological evidence.

Generative models provide a natural framework for investigating this failure mode because adversarial search can be mediated by a learned image distribution rather than performed independently over pixels. GAN-based adversarial approaches have demonstrated the feasibility of synthesizing coherent adversarial samples \cite{abc5}, but GANs can suffer from training instability, limited diversity, and image artifacts. Diffusion models provide a particularly compelling alternative because of their strong synthesis, reconstruction, and conditioning capabilities \cite{Ho2020,Rombach2022}. Rather than directly adding an optimized perturbation to an observed image, a diffusion-guided attack can optimize a representation that controls the generative process and reconstruct the adversarial image through a learned prior. The resulting search space is therefore fundamentally different: conventional attacks ask \textit{how little must the pixels change to cross the classifier boundary?}, whereas diffusion-guided attacks ask \textit{whether that boundary can be crossed while remaining constrained by a learned image-generation process}.

This distinction makes diffusion-based attacks particularly relevant to medical imaging. CXR appearance naturally varies with patient positioning, acquisition protocol, detector characteristics, image processing, and anatomical differences that is poorly represented by simple additive noise. A learned generative prior provides a mechanism for producing structured changes while retaining coherent image content. Consequently, diffusion-guided attacks can probe whether classifier predictions remain stable under plausible generative variation and can expose reliance on subtle textures, acquisition-dependent signals, spatial correlations, or other non-causal features that may not affect clinician interpretation.

Diffusion-based adversarial attacks in medical imaging, however, remain relatively underexplored. Recent text-conditioned approaches have demonstrated realistic adversarial generation and semantically meaningful pathological variation in breast ultrasound imaging \cite{Medghalchi2024}; however, this work was evaluated in a binary classification setting and did not investigate adversarial vulnerability in multi-class or multi-label disease prediction. This represents an important gap for chest radiography, where automated systems commonly evaluate multiple, potentially co-occurring thoracic abnormalities and an adversarial modification may simultaneously affect predictions across several disease classes. Language-guided latent diffusion methods such as Instruct2Attack have similarly demonstrated semantic adversarial manipulation, but primarily on natural-image benchmarks rather than clinically constrained medical imaging tasks \cite{Liu2023}. More clinically grounded approaches have also begun to emerge. The Concept-based Report Perturbation Attack (CoRPA), for example, modifies clinical concepts in radiology reports and uses text-to-image generation to construct corresponding adversarial images \cite{Rafferty2025}. Although this moves beyond purely numerical perturbations, it depends on paired radiology reports and extracted or predefined clinical concepts, potentially limiting its use for datasets containing only images and disease labels \cite{Rafferty2025}. Generative modeling has additionally been explored defensively: the Synthetic-Augmented Adversarial Training (SADA) framework combines FGSM-based adversarial training with images generated using fine-tuned Stable Diffusion and reports improved robustness and generalization \cite{SADA2024}. This work, however, addresses adversarial mitigation rather than the construction of a generative threat model.

The distinction between pixel-space and generative attacks also has implications for defense. Conventional perturbations often contain high-frequency components that can be attenuated by smoothing, compression, or denoising. Diffusion-generated adversarial images instead arise from a learned image-generation process and may contain coherent structural or textural modifications rather than separable additive noise. Defenses developed primarily around conventional perturbation patterns may therefore not generalize to this threat model. This question is increasingly relevant as CXR AI becomes integrated into triage, worklist prioritization, diagnostic support, secondary review, and downstream analytics, where inconspicuous modifications capable of changing automated disease scores could influence clinical workflows.

Importantly, attack success alone is insufficient to establish a clinically meaningful vulnerability. An adversarial image can trivially change a classifier prediction if the modification also changes the underlying disease appearance. A stronger criterion is that the machine prediction changes while the clinically relevant interpretation remains stable. Standard perceptual measures such as SSIM, LPIPS, or FID can quantify structural or distributional similarity but cannot establish preservation of diagnostic content. Clinician assessment is therefore critical for determining whether an adversarial example exposes a genuine discrepancy between machine and human interpretation rather than simply synthesizing a different pathological image.

Motivated by these considerations, we develop a diffusion-guided adversarial framework for chest X-ray classification that performs adversarial optimization through learned text conditioning rather than direct pixel-space manipulation. A learnable placeholder token is incorporated into a simple text prompt and passed through a trainable text encoder. The resulting representation conditions a frozen Stable Diffusion generator that reconstructs the input radiograph, while a frozen CXR disease classifier supplies the adversarial objective. Gradients propagate through the text-conditioning pathway, allowing the conditioning representation to be optimized toward classifier failure while image generation remains mediated by the pretrained diffusion process. Unlike report-dependent semantic attacks, the framework requires neither paired radiology reports nor manually specified clinical concepts and can therefore operate on datasets containing images and disease labels alone.

We evaluate the proposed framework across multiple CXR classifier architectures and compare it with established image-space attacks including FGSM, PGD, and C\&W. Beyond attack effectiveness, we assess image fidelity using complementary structural, perceptual, and distributional measures and incorporate clinician assessment to determine whether successful adversarial examples preserve the clinical interpretation of the corresponding radiographs. Finally, we evaluate preprocessing-based defenses and adversarial training to determine whether robustness strategies developed primarily for conventional pixel-space attacks generalize to diffusion-guided adversarial examples.

The principal contributions of this work are as follows:
\begin{itemize}

\item We introduce a diffusion-guided adversarial framework for chest X-ray classification in which adversarial optimization is performed through learnable text conditioning rather than direct pixel-level perturbation, while the diffusion generator and target disease classifier remain frozen.

\item We formulate diffusion-guided adversarial generation as a distinct robustness paradigm that searches for classifier failures through a learned generative prior, thereby probing model stability under plausible radiographic variation rather than only sensitivity to additive image-space noise.

\item We evaluate attack effectiveness across multiple classifier architectures and both binary and multi-disease CXR classification settings and compare the proposed framework with established adversarial attacks including FGSM, PGD, and C\&W.

\item We jointly evaluate adversarial effectiveness and image fidelity using structural, perceptual, and distributional measures, together with qualitative radiographic comparison, to characterize the trade-off between classifier degradation and preservation of image appearance.

\item We incorporate clinician evaluation to determine whether successful  attacks alter human diagnostic interpretation, directly examining whether changes in machine predictions occur despite preservation of clinically relevant image content.

\item We evaluate preprocessing-based defenses and adversarial training and investigate whether robustness strategies developed primarily for conventional pixel-space attacks remain effective against diffusion-guided adversarial examples.

\end{itemize}

\section{Methods}
\subsection{Problem Setup}
Let $\mathcal{D}=\{(x_i,y_i)\}_{i=1}^{N}$ denote a dataset of chest radiographs, where
$x_i \in \mathbb{R}^{H \times W \times 3}$ represents an input chest X-ray image and $y_i$ denotes the associated disease label. We consider two experimental settings: (i) binary classification, where the presence or absence of a specific disease is predicted, and (ii) multi-label classification, where multiple thoracic diseases are predicted simultaneously.

We assume access to a pretrained chest X-ray disease classifier
\begin{equation}
F : \mathbb{R}^{H \times W \times 3} \rightarrow [0,1]^C,
\end{equation}
which outputs disease probabilities for $C$ pathologies. The classifier is treated as a frozen downstream model and is not modified during adversarial generation.

Our objective is to generate an adversarial image $\tilde{x}$ that alters the classifier prediction while preserving the anatomical structure of the original radiograph. Formally, given an input image $x$, we seek
\begin{equation}
\tilde{x} = \arg\min_{\tilde{x}} \mathcal{L}(F(\tilde{x}), y^{target})
\end{equation}
subject to a similarity constraint
\begin{equation}
\tilde{x} \approx x.
\end{equation}
Unlike conventional pixel-based adversarial attacks, we enforce this constraint by operating within the latent manifold of a pretrained diffusion model.

\subsection{Architecture Overview}
\begin{figure}[t]
\centering
\includegraphics[width=\linewidth]{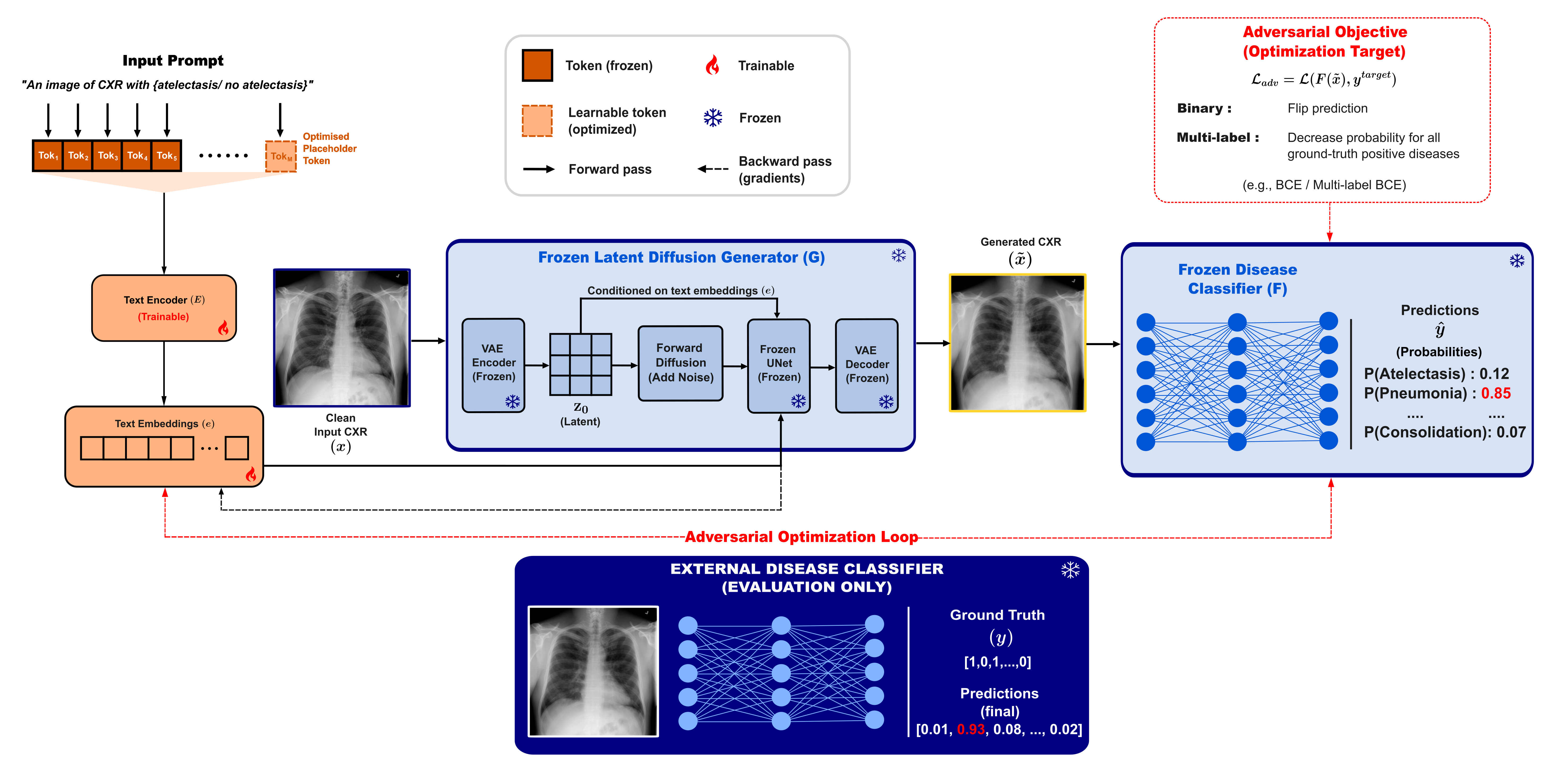}
\caption{Basic architecture diagram of the proposed diffusion-guided adversarial framework. A learnable placeholder token is passed through a trainable text encoder to generate conditioning embeddings. These embeddings guide a frozen diffusion generator that reconstructs a chest X-ray image. The generated image is evaluated by a frozen disease classifier. Gradients are propagated only through the text encoder. The diagram illustrates the binary setting; the formulation extends naturally to multi-class and multi-label classification.}
\label{fig:architecture}
\end{figure}

The overall framework is illustrated in Fig.~\ref{fig:architecture}. The architecture consists of three components:
\begin{enumerate}
\item Trainable text encoder
\item Frozen Stable Diffusion generator
\item Frozen chest X-ray disease classifier
\end{enumerate}

A learnable placeholder token $t$ is inserted into a prompt
\begin{equation}
p = \text{``chest X-ray image of } t \text{ pathology''}.
\end{equation}
The text encoder maps the prompt to an embedding
\begin{equation}
e = E(p),
\end{equation}
which conditions the diffusion generator. The generator produces a reconstructed image
\begin{equation}
\tilde{x} = G(x,e).
\end{equation}
The generated image is then evaluated by the downstream classifier
\begin{equation}
\hat{y} = F(\tilde{x}).
\end{equation}
Importantly, gradients from the adversarial objective propagate only through the text encoder, while both the diffusion model and classifier remain frozen. This ensures that adversarial perturbations are induced solely through prompt conditioning.

\subsection{Diffusion-Based Image Reconstruction}
Instead of generating images from random noise, we perform latent reconstruction to preserve anatomical realism. The input image is first encoded into latent space using a variational autoencoder (VAE)
\begin{equation}
z_0 = \text{Enc}(x).
\end{equation}
Noise is added using the forward diffusion process
\begin{equation}
z_t = \sqrt{\alpha_t} z_0 + \sqrt{1-\alpha_t}\epsilon,
\end{equation}
where $\epsilon \sim \mathcal{N}(0,I)$ and $\alpha_t$ controls the noise level at timestep $t$.
The frozen UNet predicts the noise conditioned on the text embedding
\begin{equation}
\hat{\epsilon} = \text{UNet}(z_t,t,e).
\end{equation}
The denoised latent estimate is computed as
\begin{equation}
\hat{z}_0 = \frac{1}{\sqrt{\alpha_t}}\left(z_t - \sqrt{1-\alpha_t}\hat{\epsilon}\right).
\end{equation}
Finally, the adversarial image is obtained by decoding
\begin{equation}
\tilde{x} = \text{Dec}(\hat{z}_0).
\end{equation}
This formulation ensures that perturbations remain on the diffusion manifold, preserving radiographic realism.

\subsection{Binary Disease Classification Attack}
In the binary setting, we focus on a single disease such as Atelectasis or Pneumonia. The classifier predicts
\begin{equation}
y \in \{0,1\},
\end{equation}
where $1$ denotes disease presence and $0$ denotes absence.
The goal is to generate an adversarial image that flips the classifier prediction. This models a clinically meaningful failure mode where a disease present in the radiograph is predicted as absent.
The flipped target label is defined as
\begin{equation}
y^{target} = 1 - y.
\end{equation}
The adversarial loss is
\begin{equation}
\mathcal{L}_{binary} =
- y^{target} \log \hat{y}
- (1-y^{target}) \log (1-\hat{y}).
\end{equation}
Minimizing this loss encourages the diffusion-generated image to induce a presence-to-absence or absence-to-presence prediction error.

\subsection{Multi-Class and Multi-Label Attack}
In the multi-label setting, the classifier predicts probabilities for 15 thoracic diseases simultaneously 
\begin{equation}
y \in \{0,1\}^C. 
\end{equation}
The attack is untargeted: the objective is to make the classifier incorrect on all diseases that were originally positive according to the ground-truth label. The classification loss is the negative binary cross-entropy between the predicted probabilities and the original ground-truth labels. This loss is combined with a multi-level perceptual loss computed using a frozen VGG19 network to encourage visual realism.

\subsection{Similarity Constraint}
To preserve anatomical fidelity, similarity between the original and adversarial image is measured using
\begin{equation}
\mathcal{L}_{sim} = \|x-\tilde{x}\|_2^2.
\end{equation}
Among successful adversarial candidates, the perturbation with minimum similarity loss is selected.

\subsection{Attack Success Criterion}
An adversarial attack is considered successful when the classifier prediction for the target disease is flipped relative to the ground truth label.

\paragraph{Binary Setting}
For binary disease classification, the classifier outputs a scalar probability $\hat{y}$ for disease presence. An attack is successful when the prediction crosses the decision threshold:
\begin{equation}
\text{Success} =
\mathbb{I}\left[(\hat{y} > 0.5) \neq y \right],
\end{equation}
where $y \in \{0,1\}$ denotes the ground truth label. This corresponds to flipping the prediction from disease present to absent, or vice versa.

\paragraph{Multi-Label Setting}
For multi-label classification, class-specific decision thresholds are first computed on the validation set using the precision-recall curve to maximize the F1-score for each of the 15 diseases. During the attack, these per-class thresholds are used to convert sigmoid probabilities into binary predictions. An attack is considered successful when, for every disease that was originally positive according to the ground-truth label, the classifier predicts it as absent using the corresponding tuned threshold. Predictions for other diseases are allowed to vary freely.

Successful adversarial samples are retained for downstream evaluation.

\subsection{Downstream Evaluation}
The generated adversarial images are evaluated using a standard chest X-ray disease classifier. A successful attack corresponds to a misclassification relative to the original label. This evaluation quantifies the vulnerability of existing CXR classifiers to diffusion-guided adversarial perturbations.

\subsection{Adversarial Training for Robustness}
To assess a defense strategy, we perform adversarial training. Generated adversarial samples are incorporated into the training dataset
\begin{equation}
\mathcal{D}_{aug} = \{(x_i,y_i)\} \cup \{(\tilde{x}_i,y_i)\}.
\end{equation}
The classifier is retrained on this augmented dataset to improve robustness against diffusion-based perturbations.

\subsection{Multi-Class and Multi-Label Attack}
In the multi-label setting, the classifier predicts multiple thoracic diseases simultaneously
\begin{equation}
y \in \{0,1\}^{C}.
\end{equation}
The attack is untargeted. The objective is to make the classifier incorrect on all diseases that were originally positive according to the ground-truth label. The classification loss is the negative binary cross-entropy between the predicted probabilities and the original ground-truth labels:
\begin{equation}
\mathcal{L}_{\text{cls}} = -\sum_{c=1}^{C} \Bigl[ y_c \log \hat{y}_c + (1 - y_c) \log (1 - \hat{y}_c) \Bigr].
\end{equation}
This loss is combined with a multi-level perceptual loss computed using a frozen VGG19 network to encourage visual realism:
\begin{equation}
\mathcal{L} = -\mathcal{L}_{\text{cls}} + \mathcal{L}_{\text{perc}}.
\end{equation}

\section{Results}


\begin{table*}[t]
\centering
\caption{AUROC of chest X-ray classifiers under different adversarial attacks for binary atelectasis detection and multiclass disease classification. Lower AUROC indicates greater degradation in diagnostic performance.}
\label{tab:attack_results}

\resizebox{0.8\textwidth}{!}{%
\begin{tabular}{llccc}
\toprule
Task & Attack & DenseNet & ResNet & EfficientNet\\
\midrule

\multirow{5}{*}{Binary Atelectasis}
& Clean & 0.7931 & 0.7561 & 0.6777\\
& FGSM & 0.4986 & 0.7327 & 0.6494\\
& PGD  & 0.4531 & 0.7451 & 0.6625\\
& CW   & 0.4496 & 0.6894 & 0.5337\\
& \textbf{Ours}  & \textbf{0.3885} & \textbf{0.5646} & \textbf{0.4032}\\

\midrule

\multirow{5}{*}{Multiclass}
& Clean & 0.8319 & 0.7810 & 0.7989\\
& FGSM & 0.5230 & 0.5164 & 0.5232\\
& PGD  & 0.5641 & 0.5027 & 0.4959\\
& CW   & 0.5373 & 0.5380 & 0.5080\\
& \textbf{Ours}  & \textbf{0.4878} & \textbf{0.4531} & \textbf{0.4441}\\

\bottomrule
\end{tabular}}
\end{table*}

Table~\ref{tab:attack_results} summarizes classifier performance under the evaluated adversarial attacks. Across both binary atelectasis detection and multiclass disease classification, all attacks reduced AUROC relative to clean images, demonstrating consistent adversarial vulnerability across architectures. The proposed diffusion-based attack produced the greatest overall degradation.

For binary atelectasis detection, DenseNet decreased from a clean AUROC of 0.7931 to 0.3885 under the proposed attack, while EfficientNet decreased from 0.6777 to 0.4032. ResNet was comparatively more robust but still exhibited a substantial decline from 0.7561 to 0.5646. Although FGSM, PGD, and CW also degraded performance, the proposed diffusion-based attack yielded the lowest AUROC across all three architectures.

A similar pattern was observed in the multiclass setting. The proposed attack reduced AUROC from 0.8319 to 0.4878 for DenseNet, from 0.7810 to 0.4531 for ResNet, and from 0.7989 to 0.4441 for EfficientNet. FGSM, PGD, and CW also substantially degraded performance, with several classifier--attack combinations approaching chance-level discrimination. Importantly, the consistently low AUROC produced by the diffusion-based attack across architectures demonstrates that its effectiveness extends beyond a single binary disease decision to the more complex setting of simultaneous thoracic disease classification.


\begin{table}[t]
\centering
\caption{Perceptual quality of adversarial chest X-ray images averaged across tasks. Higher SSIM indicates greater structural similarity to the original image, whereas lower LPIPS and FID correspond to improved perceptual realism.}
\label{tab:image_quality}

\setlength{\tabcolsep}{6pt}
\renewcommand{\arraystretch}{1.15}

\begin{tabular}{lccc}
\toprule
Attack & SSIM$\uparrow$ & LPIPS$\downarrow$ & FID$\downarrow$\\
\midrule
FGSM & 0.5160 & 0.8054 & 145.55\\
PGD  & 0.8124 & 0.6083 & 86.24\\
CW   & 0.8488 & 0.2397 & 62.58\\
\textbf{Ours}  & \textbf{0.9080} & \textbf{0.1670} & \textbf{51.23}\\
\bottomrule
\end{tabular}
\end{table}

Table~\ref{tab:image_quality} compares the visual fidelity of adversarial images generated by the different attacks. The proposed diffusion-based method achieved the highest structural similarity (SSIM = 0.9080) and the lowest perceptual and distributional distances (LPIPS = 0.1670; FID = 51.23). CW yielded the next-best perceptual-quality measurements, whereas PGD and particularly FGSM produced substantially greater distortion. Notably, the stronger degradation in classifier performance achieved by the diffusion-based attack was not accompanied by reduced image fidelity. Instead, the proposed method simultaneously produced the greatest overall classifier degradation and the highest similarity to the original radiographs.


\begin{figure*}[h!]
\centering
\includegraphics[width=0.7\textwidth]{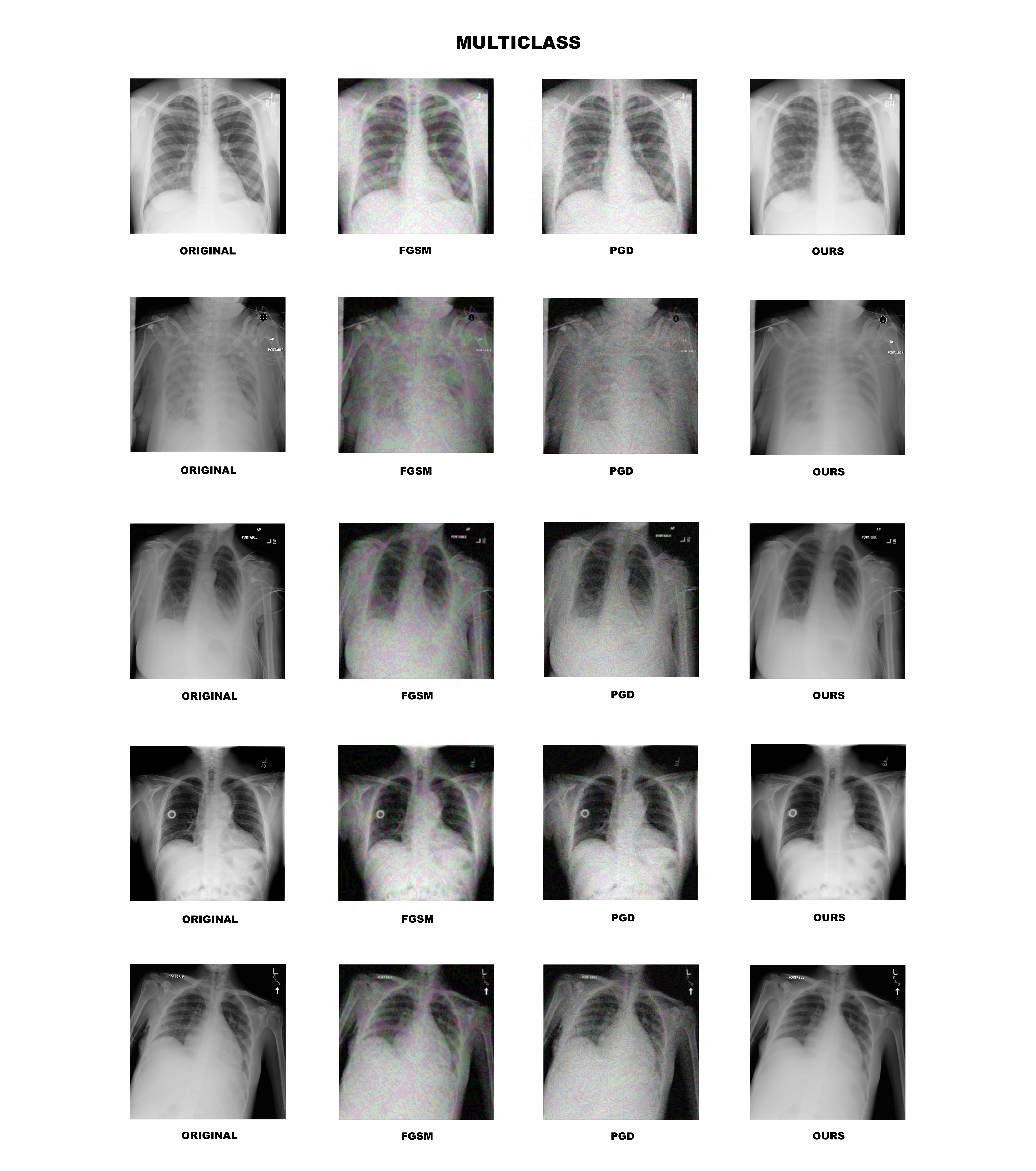}
\caption{Representative adversarial chest X-rays from the multiclass classification setting. Each row corresponds to one input radiograph and shows the original image together with adversarial examples generated using FGSM, PGD, and the proposed diffusion-based method. FGSM introduces pronounced high-frequency perturbations throughout the image, while PGD produces weaker but still perceptible perturbation patterns. In contrast, the diffusion-based method preserves the overall radiographic appearance and major anatomical structures of the source image while altering classifier predictions.}
\label{fig:multiclass_examples}
\end{figure*}


\begin{figure*}[h!]
\centering
\includegraphics[width=0.7\textwidth]{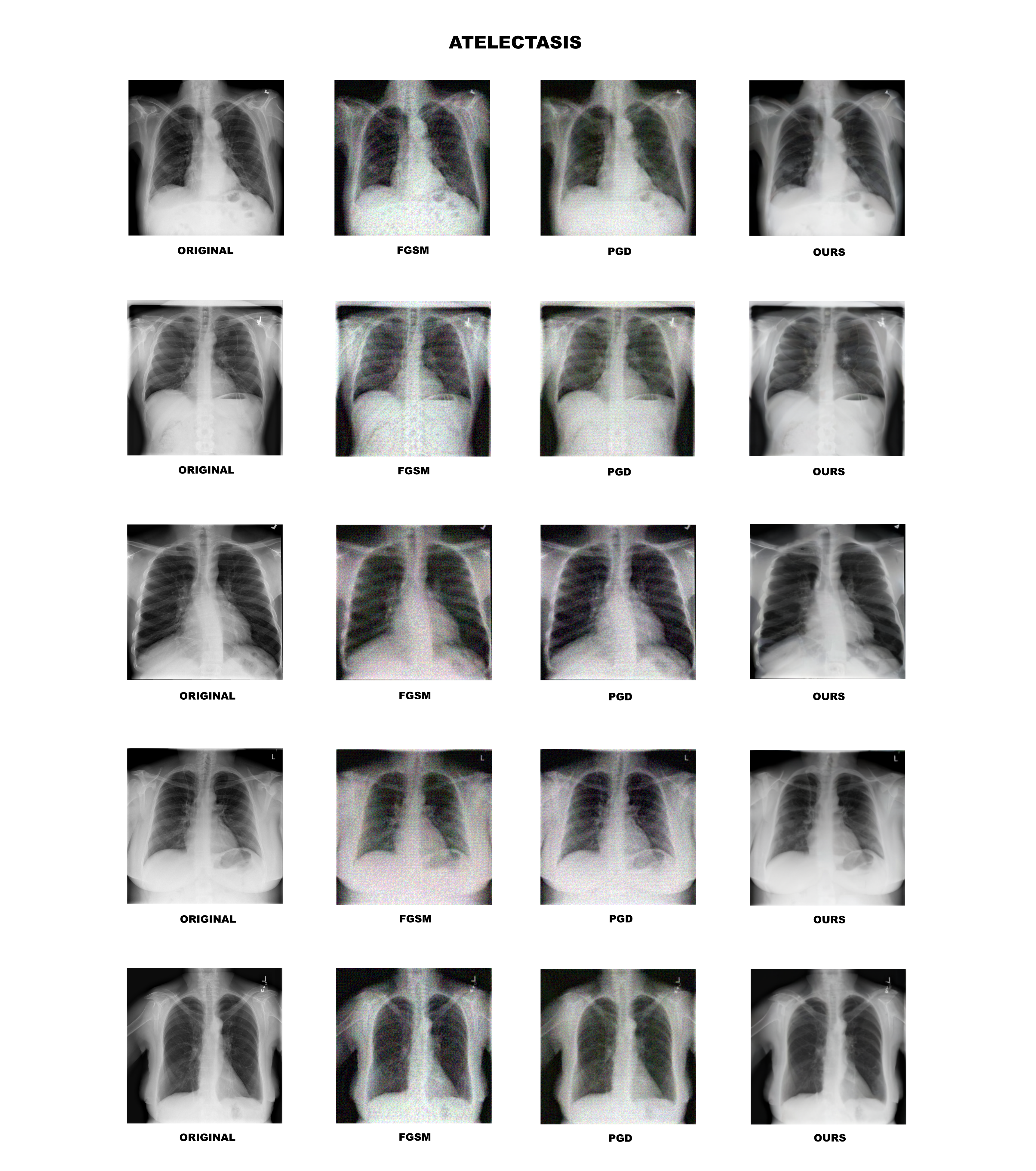}
\caption{Representative adversarial chest X-rays from the binary atelectasis classification setting. Each row shows an original radiograph and corresponding FGSM-, PGD-, and diffusion-generated adversarial images. Conventional pixel-space attacks introduce visible image-wide perturbations, particularly for FGSM, whereas diffusion-generated images remain visually similar to the source radiographs, with preservation of major thoracic structures and substantially less apparent distortion.}
\label{fig:atelectasis_examples}
\end{figure*}

The qualitative examples in Figs.~\ref{fig:multiclass_examples} and~\ref{fig:atelectasis_examples} further illustrate the distinct perceptual characteristics of the evaluated attacks. Across both classification settings, FGSM produces conspicuous high-frequency noise distributed throughout the radiograph, whereas PGD introduces less pronounced but still visible perturbation patterns, particularly within the lung fields and mediastinal regions. In contrast, diffusion-generated examples preserve global thoracic anatomy, lung morphology, the cardiac silhouette, and surrounding structures with substantially less apparent image corruption. These observations are consistent with the quantitative measurements in Table~\ref{tab:image_quality}: the attack producing the greatest overall degradation in classifier performance also most closely preserves the appearance of the source radiographs.

The representative cases in Fig.~\ref{fig:multiclass_examples} further illustrate the range of diagnostic errors induced by the attacks. In the first case, an image with an original label of Effusion was misclassified as No Finding by both FGSM and PGD, whereas the diffusion-based attack changed the prediction to Emphysema. In the second case, an image labeled Infiltration was misclassified as No Finding by all three attacks. These examples represent two distinct forms of adversarial failure: suppression of an existing abnormality, in which a pathological image is reassigned to No Finding, and diagnostic substitution, in which the predicted abnormality shifts from the original disease to a different pathological category. Thus, in a multiclass setting, successful adversarial manipulation can alter not only whether an examination is considered abnormal but also the specific disease interpretation assigned by the model.

The remaining three representative radiographs were originally labeled No Finding and demonstrate the complementary failure mode of inducing false-positive disease predictions. FGSM preserved the correct prediction in all three cases, whereas PGD changed the predictions to Atelectasis/Effusion, Pneumothorax/Emphysema, and Cardiomegaly/Effusion, respectively. The diffusion-based attack changed the corresponding predictions to Consolidation, Effusion, and Cardiomegaly. Collectively, these examples demonstrate that adversarial manipulation can produce errors in both clinically relevant directions: existing abnormalities can be suppressed or replaced by alternative disease predictions, while radiographs originally labeled as normal can acquire false-positive pathological findings. This behavior is particularly important in the multiclass setting because the consequence of an attack is not restricted to a binary label flip; rather, the inferred disease profile itself can be redirected toward a different diagnostic category. Together with the high perceptual fidelity of the diffusion-generated images, these examples demonstrate that substantial changes in model interpretation can occur without correspondingly conspicuous changes in radiographic appearance.


\begin{table}[t]
\centering
\caption{AUROC after applying different defense strategies against adversarial attacks.}
\label{tab:defense_auc}

\begin{tabular}{lcccc}
\toprule
Defense Strategy & FGSM & PGD & CW & \textbf{Ours}\\
\midrule
None & 0.5232 & 0.4959 & 0.5080 & 0.4441\\
Gaussian Smoothing & 0.5360 & 0.5118 & 0.5120 & 0.4578\\
JPEG Compression & 0.5246 & 0.5329 & 0.5247 & 0.4646\\
Adversarial Training (50\%) & 0.5681 & 0.5769 & 0.5749 & 0.4937\\
Adversarial Training (75\%) & \textbf{0.5813} & \textbf{0.5985} & \textbf{0.5991} & \textbf{0.5362}\\
\bottomrule
\end{tabular}
\end{table}

Table~\ref{tab:defense_auc} evaluates the effectiveness of the different defense strategies. Gaussian smoothing and JPEG compression provided only modest improvements over the undefended baseline. For the diffusion-based attack, AUROC increased from 0.4441 without defense to 0.4578 with Gaussian smoothing and 0.4646 with JPEG compression, indicating limited protection from conventional preprocessing. Adversarial training provided substantially greater robustness, increasing AUROC to 0.4937 with 50\% adversarial samples and to 0.5362 with 75\%. Increasing the proportion of adversarial training samples similarly improved robustness against the conventional attacks. Nevertheless, none of the evaluated defenses restored performance to the clean-image baseline, and the diffusion-based attack remained the most challenging attack under the strongest evaluated defense.


\begin{figure}[t]
\centering
\includegraphics[width=0.5\linewidth]{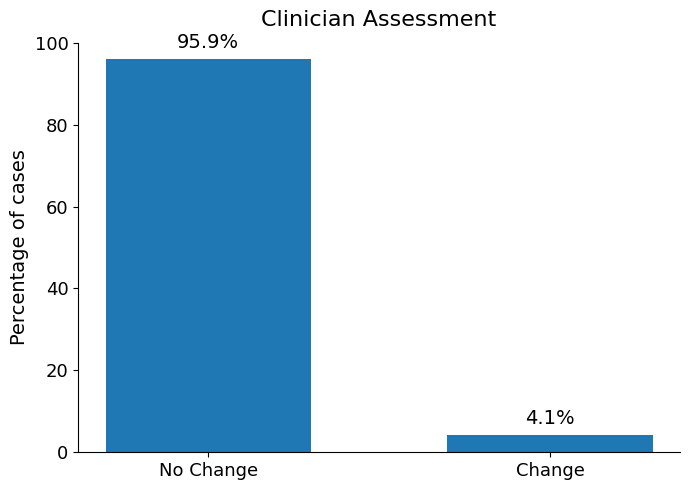}
\caption{Clinician assessment of diffusion-generated adversarial chest radiographs. Percentage of cases exhibiting unchanged or changed clinical interpretation following adversarial perturbation. The high proportion of unchanged assessments indicates that the generated adversarial modifications largely preserved clinician-perceived diagnostic content.}
\label{fig:clinician}
\end{figure}

Figure~\ref{fig:clinician} summarizes clinician assessment of the diffusion-generated adversarial chest radiographs. Despite substantial degradation in classifier performance, the clinician's diagnostic interpretation remained unchanged in 95.9\% of evaluated cases, indicating that the adversarial modifications rarely produced a clinically meaningful change in the radiographic interpretation. Only 4.1\% of cases were judged to exhibit a diagnostic change following adversarial perturbation. These findings suggest that the diffusion-guided attack can substantially alter model behavior while largely preserving the diagnostic content perceived by a clinician.

Taken together, the attack-performance, perceptual-quality, qualitative, and clinician evaluations reveal a marked discrepancy between classifier sensitivity and human clinical interpretation. The diffusion-guided attack substantially altered model predictions while frequently preserving both radiographic appearance and clinician-assessed diagnostic content. Successful attacks therefore cannot be attributed solely to conspicuous image corruption or overt changes in pathology. Rather, the results indicate that classifier decision boundaries can be crossed through generative modifications that remain clinically stable under human assessment, highlighting a failure mode that may not be adequately characterized by conventional pixel-space adversarial evaluation.

\section{Discussion}

This study demonstrates that chest X-ray classification systems can remain highly vulnerable to adversarial manipulation even when the resulting images preserve substantial visual and clinical similarity to the original radiographs. Across both binary atelectasis detection and multiclass disease classification, all evaluated attacks degraded diagnostic performance, with the proposed diffusion-guided attack producing the largest reduction in AUROC across most architectures. Importantly, this degradation was accompanied by substantially greater perceptual fidelity than conventional gradient-based attacks, suggesting that attack strength and image realism need not represent competing objectives.

This vulnerability is particularly relevant as artificial intelligence is increasingly used to automate or augment diagnostic interpretation of medical images. In such systems, image classifiers may directly identify suspected abnormalities, assign disease probabilities, generate diagnostic alerts, or provide predictions that contribute to downstream clinical decisions. Consequently, robustness is not only a question of whether a model's numerical output can be perturbed, but whether an automated diagnostic conclusion can be changed while the underlying image remains clinically consistent. Our results demonstrate precisely this failure mode: diffusion-guided modifications can substantially alter classifier predictions while preserving the apparent diagnostic content of the radiograph.

The clinical significance of this finding extends beyond conventional robustness benchmarking. Many adversarial attacks are evaluated primarily according to whether a numerical prediction can be changed under a small pixel-space perturbation. Although such experiments are useful for characterizing model sensitivity, visible high-frequency perturbations or mathematically constrained noise patterns may not reflect the types of image modifications that would plausibly enter an automated clinical imaging pipeline. In contrast, the diffusion-guided attack operates through a generative diffusion process and produces images that remain close to the distribution of realistic chest radiographs. The resulting failure mode is therefore potentially more concerning: an image may continue to appear anatomically coherent and diagnostically consistent to a human reader while eliciting a substantially different disease prediction from an automated diagnostic system.

The qualitative and perceptual evaluations support this distinction. FGSM produced conspicuous high-frequency artifacts, while PGD also introduced visible perturbation patterns. Diffusion-generated radiographs, in contrast, retained the overall appearance of the original images, including major thoracic anatomy and global image structure. This observation was consistent with the quantitative perceptual analysis, in which our method achieved the highest SSIM and lowest LPIPS and FID among all evaluated attacks. Thus, the strongest attack in terms of classifier degradation was also the one that most closely preserved the appearance of the source radiograph.

The clinician evaluation provides an additional and clinically important perspective. For the majority of evaluated diffusion-generated images, the clinician's diagnostic interpretation remained unchanged despite substantial changes in classifier behavior. This finding indicates that the adversarial modification can alter machine interpretation without producing a corresponding change in human diagnostic assessment. In the context of AI-assisted or automated diagnosis, this discrepancy is particularly important: the same radiograph may support a stable clinical interpretation while an AI system produces a substantially different diagnostic output after adversarial modification. The attack may therefore exploit features or decision boundaries relied upon by the classifier that are not equivalent to the clinically meaningful features used by human readers.

From a clinical AI safety perspective, this represents an important vulnerability as automated image interpretation becomes increasingly integrated into diagnostic workflows. AI systems may be used to detect abnormalities, prioritize suspected disease, support differential diagnosis, or provide automated assessments before or alongside radiologist review. An adversarially manipulated radiograph that remains clinically plausible but changes the output of such a system could result in a missed AI detection, an inappropriate disease alert, or an incorrect diagnostic prediction. In systems that automate portions of image interpretation, such failures could directly propagate into downstream clinical decisions. Even when a radiologist retains responsibility for the final diagnosis, incorrect upstream AI outputs could alter workflow prioritization, direct clinical attention toward or away from particular findings, or influence subsequent decision support. These findings therefore reinforce the importance of human oversight and human-in-the-loop deployment, particularly when automated predictions are used in high-stakes diagnostic settings.

The same vulnerability also has implications for malicious manipulation of medical imaging systems. A realistic adversarial attack could theoretically be used to suppress an automated detection of disease, generate false-positive AI alerts, or intentionally alter automated risk estimates without introducing obvious image corruption. Such scenarios could arise if an attacker were able to manipulate imaging data before model inference, compromise image-processing infrastructure, or modify data transmitted between systems. The relevance of this threat may increase as AI-based analysis becomes more deeply integrated with PACS, automated diagnostic and triage systems, and cloud-based inference services. Our results therefore highlight the importance of considering not only whether adversarial perturbations can fool a classifier, but whether they can alter an automated diagnostic output while remaining sufficiently realistic to escape routine visual inspection.

The findings also have implications for model validation and regulatory evaluation. Robustness assessment based only on conventional gradient-based attacks may underestimate clinically relevant vulnerabilities, particularly for AI systems intended to automate or substantially influence diagnostic interpretation. An AI system may appear resistant to readily detectable pixel perturbations while remaining vulnerable to semantic or generative attacks that operate within the data manifold. Accordingly, adversarial evaluation for medical imaging systems may benefit from incorporating generative and clinically constrained attacks in addition to established pixel-space methods. Such evaluation may be particularly important for high-risk use cases in which AI-generated predictions directly influence diagnosis, prioritization, or treatment decisions. Evaluation should therefore consider not only predictive performance under attack, but also whether discrepancies arise between automated predictions and clinically preserved image content.

The defense experiments further demonstrate that simple input preprocessing is unlikely to be sufficient. Gaussian smoothing and JPEG compression yielded only modest improvements in AUROC, indicating that attacks cannot necessarily be removed through straightforward suppression of high-frequency information. This observation is especially relevant for diffusion-guided attacks, which are not restricted to conventional additive noise patterns. Adversarial training provided substantially greater robustness, with performance improving as the proportion of adversarial examples used during training increased. Nevertheless, even 75\% adversarial training did not restore performance to clean-image levels, and diffusion-generated examples remained the most difficult to mitigate. These findings suggest that defenses designed primarily around pixel-level perturbations may be inadequate for increasingly realistic generative attacks and that technical defenses should be complemented by appropriate safeguards and human oversight in clinically consequential applications.

Several limitations should be considered. First, the clinician evaluation represents an initial assessment of clinical realism and should be expanded to multiple radiologists with different levels of expertise, together with formal inter-reader agreement analysis. Second, although perceptual metrics such as SSIM, LPIPS, and FID provide complementary measures of image similarity, they do not directly measure preservation of clinically relevant anatomy or pathology. Future evaluation could therefore incorporate region-specific analysis, pathology localization, structured radiologist scoring, and comparisons of clinically relevant image features between original and adversarial images. Third, the current experiments evaluate a limited set of disease classifiers and architectures. Future work should examine transferability across independently trained models, institutions, acquisition devices, and datasets, particularly because successful transfer would represent a more realistic black-box threat scenario. In practice, malicious access may range from complete white-box knowledge to query-only black-box interaction. Evaluating transferability and query-based optimization would therefore provide a more complete characterization of real-world security risk. In addition, the current work focuses primarily on classification. Similar generative attacks could potentially affect other components of increasingly automated radiology pipelines, including segmentation, report generation, disease localization, image quality assessment, and multimodal vision-language systems.

Finally, the objective of adversarial research in this setting is not to facilitate malicious manipulation, but to expose failure modes before they occur in deployed clinical systems. This becomes increasingly important as AI transitions from retrospective image analysis toward systems that actively support or automate components of diagnostic interpretation. Generating clinically plausible adversarial examples provides a controlled mechanism for stress-testing such systems and identifying model behavior that would otherwise remain hidden under conventional evaluation. From this perspective, diffusion-guided adversarial generation may be particularly valuable as a safety-testing tool, enabling developers to identify unsafe decision boundaries, construct more challenging validation sets, perform adversarial training, and assess whether automated diagnostic predictions are grounded in clinically meaningful image features before deployment. Coupling such robustness evaluation with appropriate human oversight may ultimately be necessary to ensure that increasing diagnostic automation does not introduce failure modes that are difficult to detect through routine clinical inspection.
\section{Conclusion}

We presented a diffusion-guided adversarial framework for generating realistic chest X-ray images capable of substantially altering the predictions of automated disease classifiers. Across binary atelectasis detection and multiclass disease classification, adversarial attacks consistently degraded classifier performance, with our method producing the greatest overall reduction in AUROC across the evaluated architectures. At the same time, the diffusion-guided method generated the most perceptually faithful adversarial images, achieving the highest SSIM and lowest LPIPS and FID among the evaluated attacks.

Clinician evaluation further showed that the diagnostic interpretation of most diffusion-generated images remained unchanged, despite substantial changes in model behavior. These findings highlight an important gap between human and machine perception: clinically plausible radiographs can remain interpretable to a clinician while causing large failures in automated diagnostic systems. This discrepancy also underscores the importance of maintaining appropriate human oversight when AI systems are incorporated into clinical workflows, particularly when model outputs may be unreliable despite apparently valid imaging inputs.

This vulnerability has direct implications for the safe deployment of medical imaging AI. As automated classifiers increasingly influence triage, prioritization, and clinical decision support, robustness evaluation must extend beyond conventional pixel-space perturbations toward realistic generative threat models. Although adversarial training improved robustness, none of the evaluated defenses completely recovered clean-image performance, indicating that clinically realistic adversarial attacks remain a challenging problem. These results further motivate human-in-the-loop deployment strategies in which automated predictions complement, rather than replace, expert clinical judgment, particularly for high-stakes decisions or cases associated with model uncertainty.

More broadly, diffusion-guided adversarial generation provides a useful framework for proactively stress-testing medical AI systems. Incorporating such attacks into model development and validation may help identify unsafe decision boundaries, improve adversarial robustness, and inform mechanisms for determining when human review should be prioritized. Combining rigorous adversarial evaluation with effective human oversight may ultimately support the development and deployment of diagnostic AI systems that remain reliable under both accidental distribution shifts and intentional manipulation.
\bibliographystyle{elsarticle-num}
\bibliography{references}
\end{document}